\documentclass[10pt,twocolumn,twoside]{IEEEtran}   

\usepackage[utf8]{inputenc}

\usepackage{macros_markus}      
\usepackage{amsmath}            
\usepackage{amsfonts}           
\usepackage{amssymb,amsthm}     
\usepackage{mathrsfs}
\usepackage{yhmath}
\usepackage{multirow}
\usepackage{stfloats}
\theoremstyle{plain}
\newtheorem{rem}{Remark}

\usepackage{graphicx}
\usepackage{epstopdf}
\usepackage[tight,footnotesize]{subfigure}
\graphicspath{{./images/}{./images/chap1/}}        

\usepackage[T1]{fontenc}
\usepackage{color}

\usepackage{algorithm}
\usepackage{algorithmicx}

\usepackage{cite}
\usepackage{balance}

\begin{document}
%
%

\title{Do Proportionate Algorithms Exploit Sparsity?}

%
\author{ Markus~V.~S.~Lima,~\IEEEmembership{Member,~IEEE,}
         Gabriel~S.~Chaves,~\IEEEmembership{Student Member,~IEEE,} \\
         Tadeu~N.~Ferreira,~\IEEEmembership{Member,~IEEE,}
         Paulo~S.~R.~Diniz,~\IEEEmembership{Fellow,~IEEE,}
\thanks{M.V.S.~Lima, G.S.~Chaves, and P.S.R.~Diniz are with the PEE/COPPE, Universidade Federal do Rio de Janeiro (UFRJ),
Rio de Janeiro, RJ, 68504, Brazil (e-mails: \{markus.lima,gabriel.chaves,diniz\}@smt.ufrj.br).}
\thanks{T.N.~Ferreira is with the Universidade Federal Fluminense (UFF), Niteroi, RJ, Brazil (e-mail: tadeu\_ferreira@id.uff.br).}
}

\markboth{}%
{Lima \MakeLowercase{\textit{et al.}}: DO PROPORTIONATE ALGORITHMS EXPLOIT SPARSITY?}
%



\maketitle

\begin{abstract}  
Adaptive filters exploiting sparsity have been a very active research field, among which the algorithms that follow the 
``proportional updates'' principle, the so-called proportionate-type algorithms, are very popular.
Indeed, there are hundreds of works on proportionate-type algorithms and, therefore, their advantages are widely known.
This paper addresses the unexplored drawbacks and limitations of using proportional updates and their practical impacts.
Our findings include the theoretical justification for the poor performance of these algorithms in several sparse scenarios, and also 
when dealing with non-stationary and compressible systems.
Simulation results corroborating the theory are presented.
\end{abstract}

\begin{IEEEkeywords}
Adaptive filtering, sparsity, proportionate.
\end{IEEEkeywords}

%
\IEEEpeerreviewmaketitle

\section{Introduction}
\label{sec:intro}

Adaptive filters exploiting sparsity have been a very active research field in the last twenty years and they find applications in many areas like  
echo cancellation~\cite{Duttweiler_PNLMS_tsap2000,Sugiyama_proportionate_icassp2004,Naylor_iipnlms_sigpro2006,Paleologu_papaEcho_spl2010}, 
channel estimation~\cite{Taheri_ZALMS_analysis_SP2009,Markus_performEval_eusipco2017}, 
system identification~\cite{Markus_sparseSMAP_tsp2014,Markus_apssi_icassp2013,Vitor_SparsityAwareAPA_sspd2011,Mariane_SparseSystemID_iswcs2010,Theodoridis_l1ball_icassp2010}, 
and modeling of nonlinear systems~\cite{Markus_l0Volterra_eusipco2019}. 
In this context, proportionate-type algorithms have received considerable attention. 
In these algorithms, the adaptation step applied to each coefficient is proportional (to a certain extent) to the magnitude of such coefficient and, 
as a result, high magnitude coefficients are learned faster.

The {\it proportionate normalized least mean square} (PNLMS) algorithm was the first of its kind~\cite{Duttweiler_PNLMS_tsap2000}.
Later, researchers have realized that applying updates proportional to the magnitude of the coefficients, although interesting in sparse scenarios, 
could lead to poor performance in non-sparse (dispersive) scenarios~\cite{Benesty_IPNLMS_icassp2002}.
Thus, many variations of the PNLMS algorithm have been proposed, and as expected, their updates have {become} ``less proportional'' to increase 
the robustness of the proportionate-type algorithms against other classes of systems. 
For instance, the update scheme of the {\it improved PNLMS} (IPNLMS) algorithm, one of the most successful variations, combines the characteristics of both the 
PNLMS and NLMS algorithms, thus mixing the fast convergence in sparse systems of the former with the robustness of the latter~\cite{Benesty_IPNLMS_icassp2002}. 
There are many other variations of the PNLMS algorithm~\cite{Liu_muLawPNLMS_icassp2008,Naylor_scipnlms_acssc2006,Naylor_iipnlms_sigpro2006}, 
and also generalizations based on the affine projection (AP) 
algorithm~\cite{Eneroth_papa_waspaa1999,Sugiyama_proportionate_icassp2004,Paleologu_papaEcho_spl2010,Diniz_sm_pap_jasmp2007}.


While the advantages of proportionate-type algorithms are widely known and have been exploited for years, there is no study addressing their limitations. 
In this paper, we explain one fundamental issue related to the {\it proportional updates principle} and some practical issues. 
For the sake of clarity, we use NLMS-based algorithms, 
but our conclusions can be extended to more general proportionate-type algorithms.  

This work is organized as follows.
In Section~\ref{sec:pt_nlms}, we provide a unified review of {\it proportionate-type NLMS} (Pt-NLMS) algorithms and cover the two most famous: 
the PNLMS and IPNLMS algorithms.
In Section~\ref{sec:properties}, we address several properties of Pt-NLMS algorithms. 
We start by establishing a connection between the NLMS and Pt-NLMS algorithms. 
Then we analyze the mean squared error (MSE) surface in order to understand their convergence characteristics and main limitation. 
Next, the practical issues due to this main limitation are discussed.
Simulation results corroborating these issues are shown in Section~\ref{sec:simulations}.
The conclusions are drawn in Section~\ref{sec:conclusion}.

{\it Notation:} Scalars are represented by lowercase letters. 
Vectors (matrices) are denoted by lowercase (uppercase) boldface letters. 
The $i$-th entry of a vector $\wbf$ is denoted by $w_i$.
At iteration $k$, $\wbf(k), \xbf(k) \in \Rset^{N+1}$ denote the adaptive filter coefficient (or weight) vector and the 
input vector, respectively, where $N$ is the filter order. 
The filter output and error signals are defined as $y(k) \triangleq \wbf^T(k) \xbf(k)$ and $e(k)\triangleq d(k)-y(k)$, 
where $d(k) \in \Rset$ is the desired/reference signal. 
The $\ell_1$ norm of a vector $\wbf\in\mathbb{R}^{N+1}$ is given by $\|\wbf\|_1=\sum_{i=0}^N|w_i|$.
$\Ibf$, ${\bf 0}$ and $\Diag{\wbf}$ stand for the identity matrix, the null vector, and the diagonal matrix having $\wbf$ on its main diagonal. 


\section{Proportionate-type NLMS Algorithms}
\label{sec:pt_nlms}

{Pt-NLMS algorithms have their update equations (recursions) given by the following general form~\cite{Markus_sparseSMAP_tsp2014}}
\begin{align} \label{eq:Pt_NLMS_recursion}
	\wbf(k+1) = \wbf(k) + \mu \frac{e(k) \Gbf(k) \xbf(k)}{\xbf^{T}(k) \Gbf(k) \xbf(k) + \delta}~,
\end{align}
in which $\Gbf(k)$ is the so-called {\it proportionate matrix}, a diagonal matrix whose elements on the main diagonal, denoted by 
$g_i(k), i =0, 1, \dots, N$, have some proportional relation with their corresponding weights $w_i(k)$.
The regularization parameter $\delta \in \Rset_+$ is a small nonnegative number used to avoid numerical issues when $\xbf^{T}(k) \Gbf(k) \xbf(k)$ 
tends to zero. 
The step size should be chosen in the range $0 < \mu \leq 1$.


The proportionate matrix performs an uneven distribution of the step size among the weights during the update process. 
That is, entries $g_i(k)$ corresponding to high magnitude weights $w_i(k)$ will have high values, thus accelerating the convergence 
of these weights to their optimal values, whereas those $g_i(k)$ corresponding to low magnitude weights will have low values, thus slowing 
their convergence rate. 
Observe that if $\Gbf(k) = \Ibf$, then all the coefficients $w_i(k)$ share the same step size during the updates 
and the NLMS recursion is obtained as a particular case of~\eqref{eq:Pt_NLMS_recursion}. 
In the following subsections, we present two possibilities for $\Gbf(k)$ which gave rise to the 
PNLMS and the IPNLMS algorithms.

\subsection{The PNLMS Algorithm}
\label{sub:pnlms}

In the PNLMS algorithm, whose recursion is described in~\eqref{eq:Pt_NLMS_recursion}, the entries of $\Gbf(k) = \Diag{[g_{0}(k) ~ g_1(k) ~ \cdots ~ g_{N}(k)]}$ 
are given by~\cite{Duttweiler_PNLMS_tsap2000}
\begin{align}
\hspace{-3mm}       g_{i}(k)      &= \dfrac{\gamma_{i}(k)}{\sum\limits_{j = 0}^{N} \lvert \gamma_{j}(k) \rvert}~, \text{ for $i=0,1,\dots,N$, and }        \label{eq:g_pnlms}   \\
\hspace{-3mm}     \gamma_{i}(k)\! &= \! \max \! \left\{ \! \underbrace{ \lvert w_{i}(k) \rvert }_{\rm 1^{st} \ term},\underbrace{ \rho\max \! \left\{ \delta_{\rm P},\lvert w_{0}(k) \rvert,\cdots \! ,\lvert w_{N}(k) \rvert \right\} }_{\rm 2^{nd} \ term} \! \right\} \!\!,   \label{eq:gamma_pnlms}
\end{align}
where $\max\{\cdot\}$ returns the maximum element of a set, 
$\delta_{\rm P} \in \mathbb{R}_+$ is used to prevent $\wbf(k)$ from stalling during the initialization stage, in case the coefficients are 
initialized as $\wbf(0) = {\bf 0}$, and its typical value is $\delta_{\rm P} = 0.01$, and 
$\rho \in \mathbb{R}_+$ is used to reduce the ${\rm 2^{nd} \ term}$ so as to prioritize the ${\rm 1^{st} \ term}$ in~\eqref{eq:gamma_pnlms} 
and its typical value is $\rho = 0.01$.
These constants were introduced to avoid the aforementioned numerical problems, which would occur if 
$g_i(k) = {\lvert w_i(k) \rvert}/{\| \wbf(k) \|_1}$,\footnote{This term represents the original concept of {\it proportional updates} and also
justifies the name of the algorithms inspired by it~\cite{Duttweiler_PNLMS_tsap2000}.}
{whenever $w_i(k)  \rightarrow  0$.}

\subsection{The IPNLMS Algorithm}
\label{sub:ipnlms}

In the IPNLMS algorithm, whose recursion is described in~\eqref{eq:Pt_NLMS_recursion}, the entries of $\Gbf(k) = \Diag{[g_{0}(k) ~ g_1(k) ~ \cdots ~ g_{N}(k)]}$ 
are given by~\cite{Benesty_IPNLMS_icassp2002}
\begin{align}
	g_{i}(k) = \underbrace{ \dfrac{1-\alpha}{2(N+1)} }_{\rm NLMS \ term} + \underbrace{ \dfrac{(1+\alpha)\lvert w_{i}(k) \rvert}{2\lVert \wbf(k) \rVert_{1} + \delta_{\rm IP}} }_{\rm PNLMS \ term}~, \label{eq:gn_ipnlms}
\end{align}
where $\delta_{\rm IP} \in \mathbb{R}_+$ is a regularization factor used to avoid numerical problems when $\| \wbf(k) \|_1$ tends to zero, and 
$\alpha \in [-1,1)$ is a real number that represents a tradeoff between the NLMS and PNLMS terms in~\eqref{eq:gn_ipnlms}. 
If $\alpha = -1$, then the IPNLMS is equivalent to the NLMS algorithm, whereas for $\alpha \approx 1$ its update resembles that of the PNLMS algorithm. 

The IPNLMS algorithm was designed to be a more robust version of the PNLMS algorithm, whose performance can be disappointing in many cases, 
like in the identification of dispersive systems~\cite{Benesty_IPNLMS_icassp2002}. 
To do so, the IPNLMS algorithm must behave more like an NLMS algorithm than like a PNLMS, which justifies the usual choice of $\alpha = -0.5$.

\section{Properties of Pt-NLMS algorithms}
\label{sec:properties}


\subsection{Connection with the NLMS algorithm}
\label{sub:connection}

As previously explained, the proportionate matrix performs an uneven distribution of the step size in order to 
accelerate the convergence of the high magnitude coefficients.
In practice, $\Gbf(k)$ is set as a function of the adaptive filter coefficients 
$\wbf(k)$, which is our best guess of the optimal (and unknown) 
coefficients $\wbf_{\rm o}$. 
Let us assume that the proportionate matrix is set based on $\wbf_{\rm o}$, thus leading to the best possible proportionate matrix 
$\Gbf_{\rm o}$.\footnote{Although unfeasible, this assumption not only facilitates the analysis by generating a deterministic 
proportionate matrix $\Gbf_{\rm o}$, but also allows us to better comprehend the algorithm dynamics, especially after some iterations when 
it is expected that $\Gbf(k)$ becomes a better approximation of $\Gbf_{\rm o}$.}
Replacing $\Gbf(k)$ with $\Gbf_{\rm o}$ and premultiplying~\eqref{eq:Pt_NLMS_recursion} by $\Gbf_{\rm o}^{-\frac{1}{2}}$ we obtain
\begin{align} \label{eq:nlms}
   \wbf'(k+1) = \wbf'(k) + \mu \frac{e(k)  \xbf'(k)}{\xbf'^{T}(k)  \xbf'(k) + \delta}~,
\end{align}
where $\wbf'(k) \triangleq \Gbf_{\rm o}^{-\frac{1}{2}} \wbf(k)$, $\xbf'(k) \triangleq \Gbf_{\rm o}^{\frac{1}{2}} \xbf(k)$ and 
the error can be written as
\begin{align} \label{eq:error_changeVariab}
 e(k) \!=\! d(k) - \wbf^T \Gbf_{\rm o}^{-\frac{1}{2}} \Gbf_{\rm o}^{\frac{1}{2}} \xbf(k) \!=\! d(k) - \wbf'^T(k) \xbf'(k).
\end{align}
The conclusion is that a Pt-NLMS algorithm with the optimal proportionate matrix can be regarded as an NLMS algorithm with weights $\wbf'(k)$, 
operating on a transformed input vector $\xbf'(k)$ so as to generate its output $y'(k) \triangleq \wbf'^T(k) \xbf'(k)$.

\subsection{MSE Surface}
\label{sub:surface_MSE}

The connection with the NLMS algorithm allows us to interpret the Pt-NLMS as a stochastic gradient 
algorithm~\cite{Diniz_adaptiveFiltering_book2020,Haykin_adaptiveFiltering_book2002}. 
In this way, the MSE surface can help us understand the convergence characteristics of this algorithm.
So, let us evaluate the MSE for some fixed (and known) values of the coefficient vector $\wbf'$
\begin{align}
 \xi  &= \Expval{e^2(k)} = \Expval{ \left( d(k) - \wbf'^T \xbf'(k) \right)^2 }           \nonumber \\
      &= \sigma_d^2 - 2 \pbf'^T \wbf' + \wbf'^T \Rbf' \wbf' ,
\end{align}
where $\sigma_d^2 \triangleq \Expval{d^2(k)}$, $\Rbf' \triangleq \Expval{\xbf'(k) \xbf'^T(k)}$ is the autocorrelation matrix of the transformed input vector $\xbf'(k)$
and $\pbf' \triangleq \Expval{d(k) \xbf'(k)}$ is the cross-correlation between input and desired signals, and we are assuming that both 
$\xbf'(k)$ and $d(k)$ are zero-mean jointly wide-sense stationary (WSS) processes. 
After some mathematical manipulations and using the fact that the 
{Wiener filter gives the minimum MSE solution} $\wbf'_{\rm o} \triangleq \Rbf'^{-1} \pbf'$, the
previous equation can be written as 
\begin{align}
 \xi  \!=\!  \xi_{\rm min} + \Delta\wbf'^T  \Rbf' \Delta\wbf'                             
      \!=\!  \xi_{\rm min} + \Delta\wbf'^T  \Gbf_{\rm o}^{\frac{1}{2}} \Rbf \Gbf_{\rm o}^{\frac{1}{2}} \Delta\wbf',    \label{eq:surface_MSE}
\end{align}
where $\xi_{\rm min} \triangleq \sigma_d^2 - \wbf'^T_{\rm o}  \Rbf' \wbf'_{\rm o}$, $\Delta\wbf' \triangleq \wbf' - \wbf'_{\rm o}$,  
$\Rbf \triangleq \Expval{\xbf(k) \xbf^T(k)}$ is the autocorrelation matrix of the input vector $\xbf(k)$, 
and $\Rbf' = \Gbf_{\rm o}^{\frac{1}{2}} \Rbf \Gbf_{\rm o}^{\frac{1}{2}}$ due to the definition of $\xbf'(k)$.

Equation~\eqref{eq:surface_MSE} states that the MSE surface of a Pt-NLMS algorithm is a hyperparaboloid in the parameter space with Hessian matrix 
determined by $\Rbf'$ instead of $\Rbf$, the Hessian matrix related to the MSE surface of the NLMS algorithm. 
Both $\Rbf'$ and $\Rbf$ are symmetric positive definite matrices and, therefore, they lead to a convex hyperparaboloid and their eigenvalues are 
real positive numbers.
The condition number of such matrices is of paramount importance as it is related to 
the shape of the hyperellipses of constant MSE (herein called {\it contours}) which, in its turn, is associated with 
the convergence speed of gradient-based algorithms.
For matrices like $\Rbf'$ and $\Rbf$, we can define the condition number as
 $\kappa(\Rbf) \triangleq {\lambda_{\rm max}(\Rbf)}/{\lambda_{\rm min}(\Rbf)} ,$
where $\lambda_{\rm max}(\Rbf)$ and $\lambda_{\rm min}(\Rbf)$ denote the maximum and minimum eigenvalues of $\Rbf$, respectively.

For instance, let us consider the case in which the parameters live in $\Rset^2$. 
If the input signal is white with $\Rbf = \sigma_x^2 \Ibf$, where $\sigma_x^2$ represents its variance, then $\kappa(\Rbf) = 1$ 
resulting in circular contours. 
As the input signal becomes more correlated, the condition number becomes larger, resulting in contours comprised of ellipses having one axis 
increasingly larger than the other. 
In the limiting case where the condition number goes to infinity, the ellipses degenerate into parallel lines. 
The shapes of these contours have a direct impact on the convergence rate of gradient-based algorithms, as the gradient direction 
at a given point is orthogonal to the tangent line passing through such point~\cite{Antoniou_optmization_book2007}.
Thus, for circular contours, fast convergence is expected since the gradient direction forms a line passing through the center of these contours 
(i.e., the optimal point).
On the other hand, as $\kappa$ increases, the line formed by the gradient direction becomes increasingly further from the center resulting in slower 
convergence.

For Pt-NLMS algorithms we have 
\begin{align}
 \kappa(\Rbf') \!=\! \kappa(\Gbf_{\rm o}^{\frac{1}{2}} \Rbf \Gbf_{\rm o}^{\frac{1}{2}}) \! \leq \! \left[ \kappa(\Gbf_{\rm o}^{\frac{1}{2}}) \right]^2 \! \kappa(\Rbf)    
                                                                                    \!=\! \kappa(\Gbf_{\rm o})    \kappa(\Rbf) ,
\end{align}
where the inequality is a property of $\kappa$ and the last equality is valid since $\Gbf_{\rm o}$ is a diagonal matrix. 
In words, the condition number of $\Rbf'$ can be up to $\kappa(\Gbf_{\rm o})$ times higher than that of $\Rbf$. 
The value of $\kappa(\Gbf_{\rm o})$ depends on the selected algorithm and also on the optimal coefficients $\wbf_{\rm o}$. 
For example, for the {\it proportional updates} principle, we have  
 $\kappa(\Gbf_{\rm o}) = {\max\limits_i |w_{{\rm o},i}|}/{\min\limits_j |w_{{\rm o},j}|} ,~\text{for}~i, j \in \{0, 1, \dots, N\},$
i.e., the ratio between the maximum and minimum magnitudes of the entries of $\wbf_{\rm o}$, which can be very large in sparse scenarios.
In summary, for sparse systems, the proportionate matrix leads to $\kappa(\Rbf') \gg \kappa(\Rbf)$, i.e., it increases the correlation of the input signal. 
Regarding the MSE surface, matrix $\Gbf_{\rm o}$ tends to generate almost parallel lines as contours (e.g., see Fig.~\ref{fig:supmse2}).
Consequently, gradient-based methods will be very slow in learning the small magnitude coefficients (those related to the largest axes of the hyperellipses).



\subsection{Practical Limitations of Pt-NLMS algorithms}
\label{sub:issues}

In the previous subsection, the {\it main issue} related to Pt-NLMS algorithms, which concerns their {\it slow convergence rate for the low magnitude 
coefficients}, was explained using a rigorous approach that {quantifies} how much harm this issue can introduce.
Here, we focus on practical limitations due to this main issue. 
They are summarized in the following three remarks.
\begin{rem}[Initialization]\label{rem:initialization}
 Pt-NLMS algorithms require that $w_i(0) \approx w_{{\rm o},i}$ for every $i$ corresponding to low magnitude entries of the optimal coefficients $\wbf_{\rm o}$.
\end{rem}
\begin{rem}[Support]\label{rem:support}
 The position of the high magnitude coefficients within $\wbf_{\rm o}$ (herein called {\it support}) matters.
 The performance of Pt-NLMS algorithms becomes worse as the support moves to the end of the vector.
 \end{rem}
\begin{rem}[Tracking]\label{rem:tracking}
 Pt-NLMS algorithms are not suitable for the tracking of time-varying coefficients $\wbf_{\rm o}(k)$ in which the support may vary during the iterations.
\end{rem}

Remark~\ref{rem:initialization} states that the weights $\wbf(0)$ must be initialized very close to the low magnitude entries of $\wbf_{\rm o}$, which  
is a straightforward consequence of the slow convergence of these coefficients. 
This is not a problem for sparse systems, since their low magnitude entries are equal to zero thus allowing the standard initialization $\wbf(0) = {\bf 0}$, 
but can be a problem when dealing with compressible systems, i.e., systems whose energy is concentrated in a few coefficients and, therefore, their low magnitude 
entries are close (but not necessarily equal) to 0.

To explain Remark~\ref{rem:support}, let us suppose that the first nonzero entry of $\wbf_{\rm o}$ is indexed by $k_0 \gg 1$. 
Assuming a tapped-delay-line structure for $\xbf(k)$~\cite{Diniz_adaptiveFiltering_book2020}, during the iterations $1 \leq k < k_0$,
the coefficients $\wbf(k)$ are updated only due to noise, since $d(k) = n(k)$ at these iterations. 
{The adaptive filter will start to learn $\wbf_{\rm o}$ only at iteration $k_0$, i.e., after several updates that lead to  
$\wbf(k_0) \neq  {\bf 0}$, meaning that the $w_i(k_0) \neq 0 = w_{{\rm o},i}$, for $i \in \{0, 1, \dots, k_0 -1 \}$, 
thus these $w_i$ will converge slowly due to the {\it main issue} stated in the beginning of this subsection.}

It is not difficult to explain Remark~\ref{rem:tracking} using Remarks~\ref{rem:initialization} and~\ref{rem:support}. 
Indeed, for time-varying coefficients $\wbf_{\rm o}(k)$, the support and the values of the optimal coefficients may change. 
For example, suppose that at a {particular} iteration, a low magnitude coefficient $w_{{\rm o},i}(k)$ changes to a high magnitude coefficient $w_{{\rm o},i}(k+1)$.
If we have a good estimate of $w_{{\rm o},i}(k)$, i.e., $w_i(k) \approx w_{{\rm o},i}(k)$, then the convergence rate related to this coefficient will 
be very slow, meaning that it will take more time to learn $w_{{\rm o},i}(k+1)$.

Finally, it is worth mentioning that these remarks are valid for any kind of $\wbf_{\rm o}$, even for sparse systems, as confirmed by the simulations 
presented in the next section.

\section{Simulations}
\label{sec:simulations}

\begin{figure*}[t]
			\vspace*{-1mm}
	\centering
	\subfigure[b][]{\includegraphics[width=.3\linewidth]{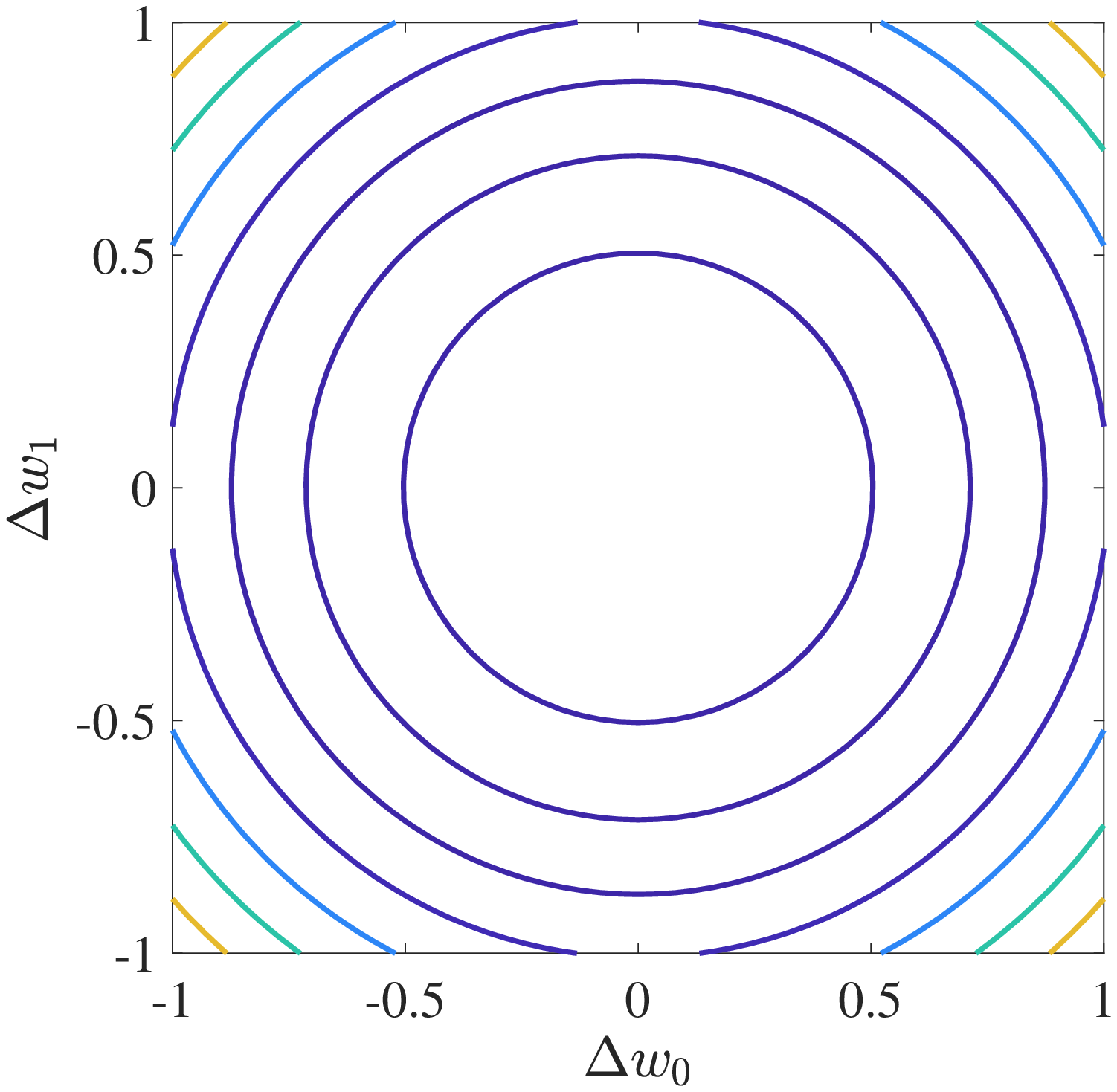}
		\label{fig:supmse1}}~
	\subfigure[b][]{\includegraphics[width=.3\linewidth]{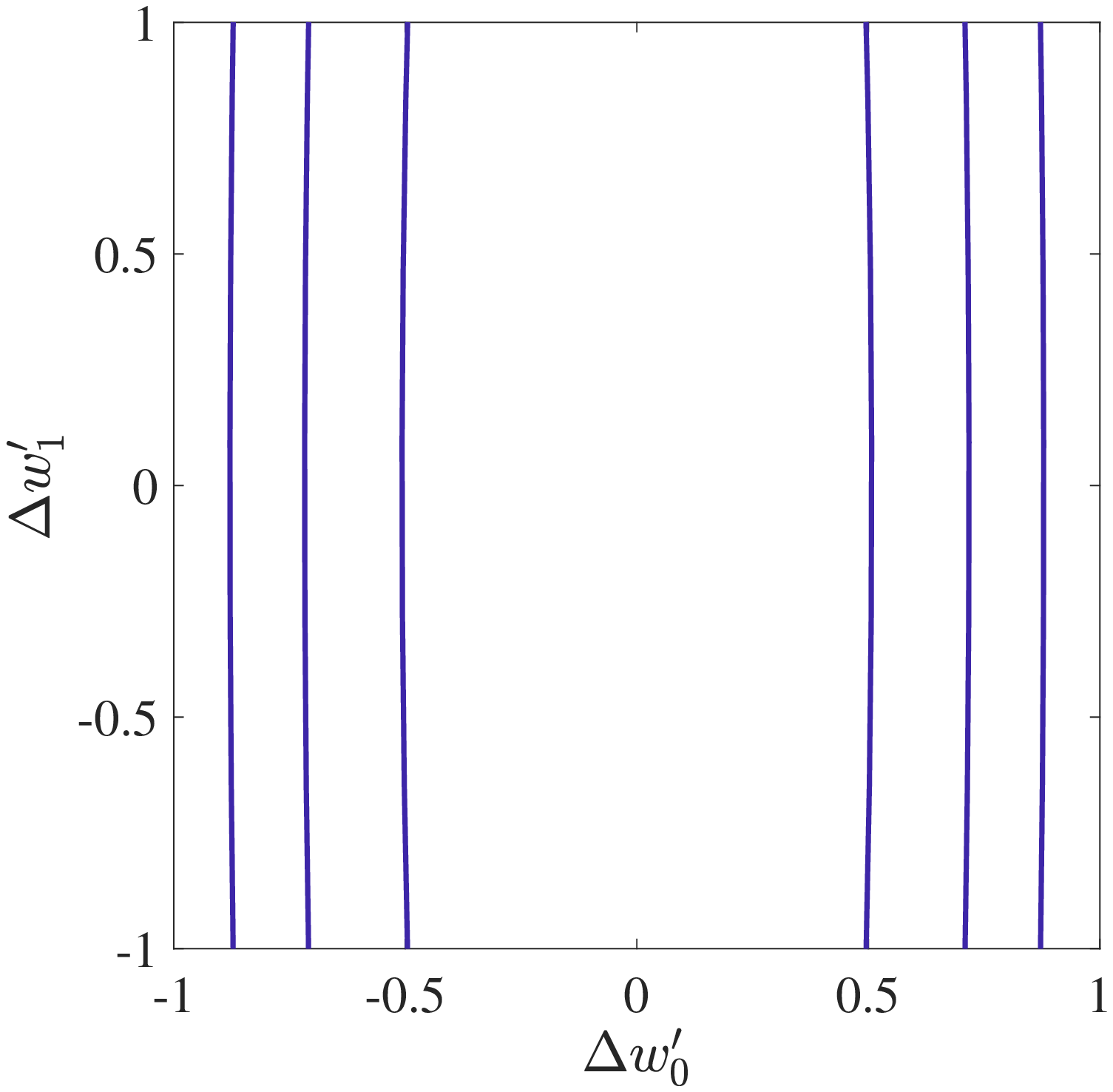}
		\label{fig:supmse2}}~
	\subfigure[b][]{\includegraphics[width=.28\linewidth]{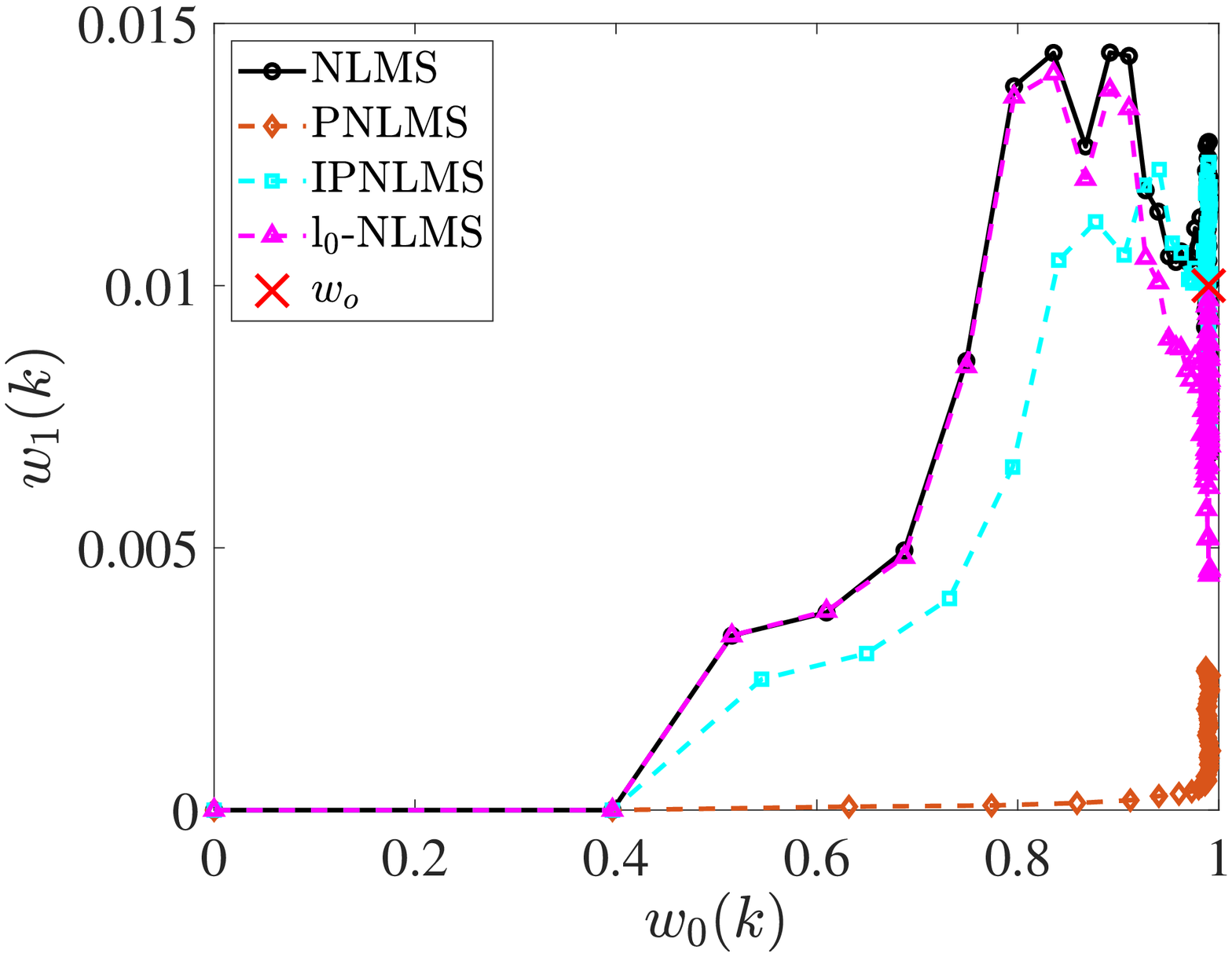}
		\label{fig:conv_path}}
	\vspace*{-0.3cm}
	\caption{Contours of the MSE surface for $\wbf_{{\rm o}}^{(0)}=[0.99~0.01]^T$: 
	(a) NLMS algorithm; (b) PNLMS algorithm; (c) Convergence path of the coefficients. \label{fig:supmse}}
		\vspace*{-0.2cm}
\end{figure*}
\begin{figure*}[t]
			\vspace*{-1mm}
	\centering
	\subfigure[b][]{\includegraphics[width=.28\linewidth]{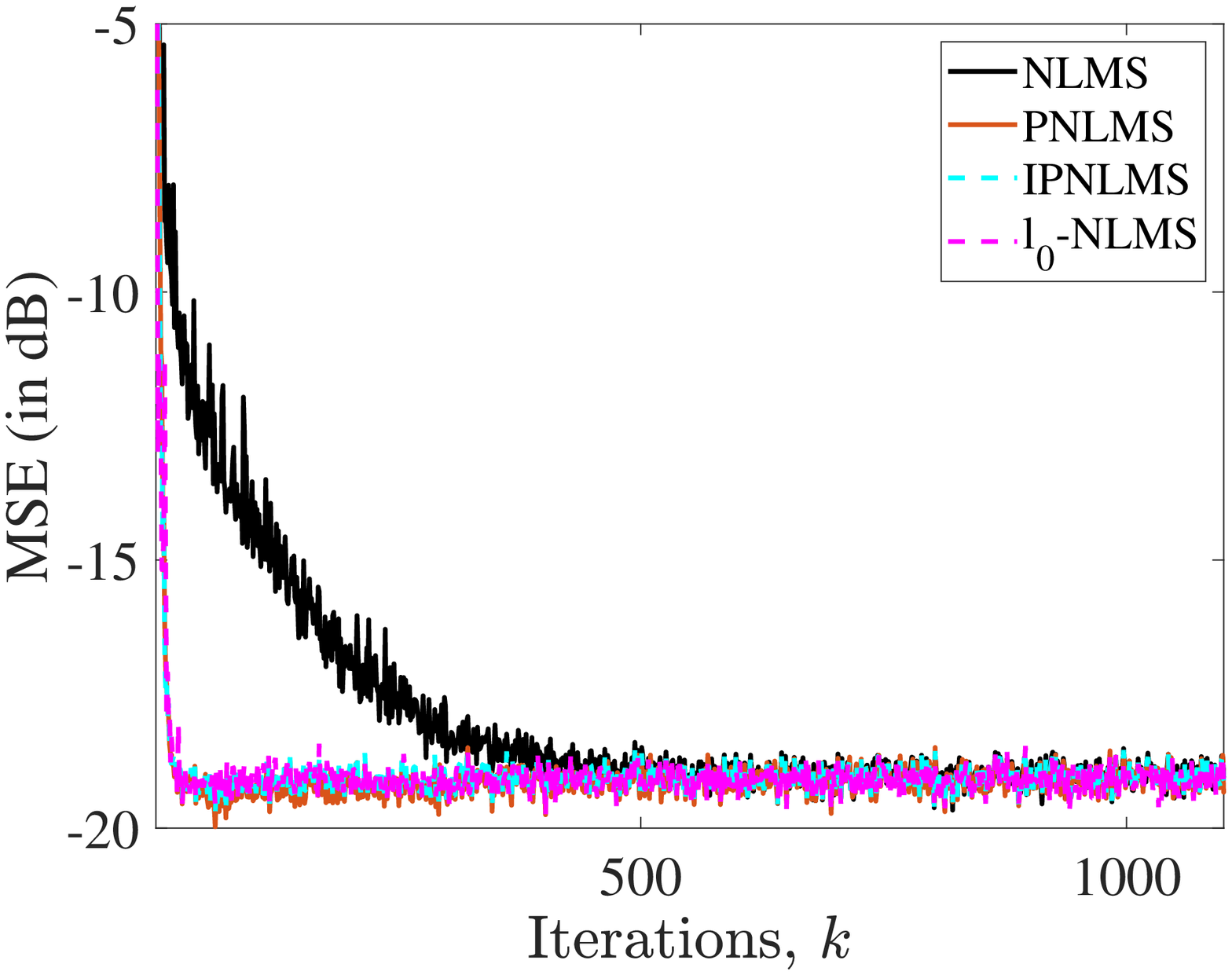}
		\label{fig:bad_init00}}~
	\subfigure[b][]{\includegraphics[width=.28\linewidth]{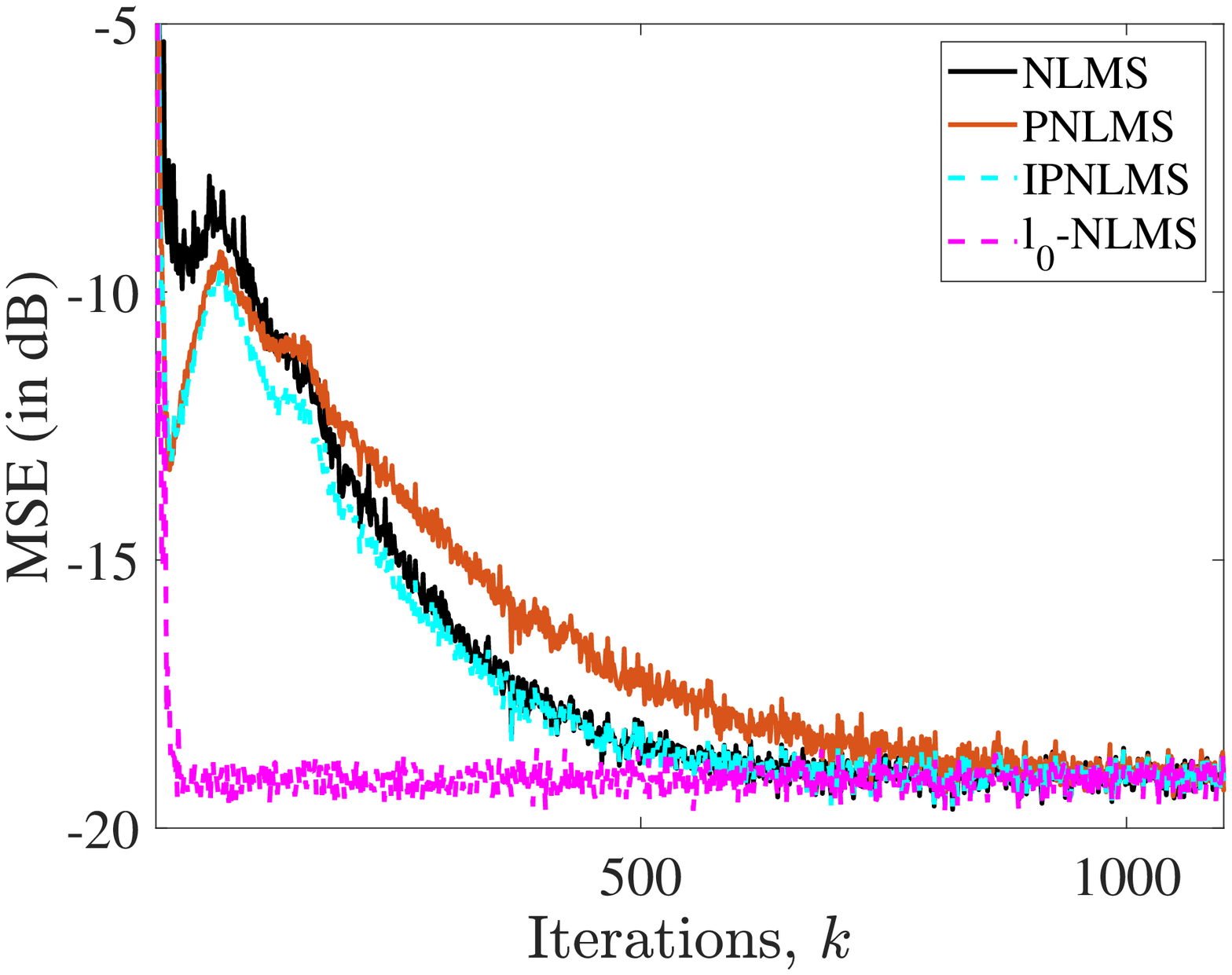}
		\label{fig:bad_init01}}~
	\subfigure[b][]{\includegraphics[width=.28\linewidth]{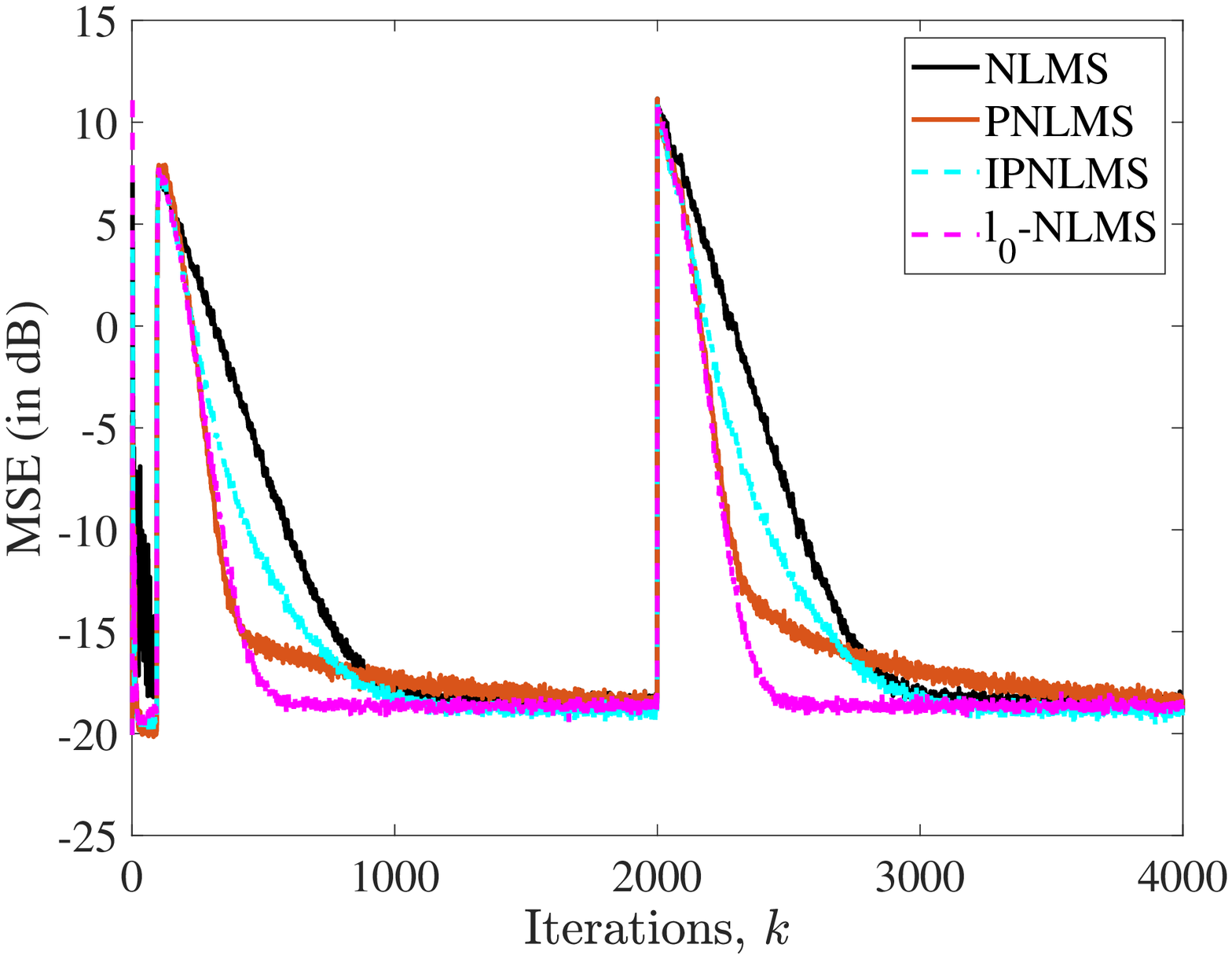}
		\label{fig:track}}
	\vspace*{-0.3cm}
	\caption{MSE learning curves considering: (a) $\wbf_{{\rm o}}^{(1)}$ with $\wbf(0)=[0~\dots~0]^T$; 
	(b) $\wbf_{{\rm o}}^{(1)}$ with $\wbf(0)=[0.05~\dots~0.05]^T$; 
	(c) the non-stationary system given in~\eqref{eq:non_stationary} with $\wbf(0)=[0~\dots~0]^T$. \label{fig:remarks}}
	      \vspace*{-0.2cm}
\end{figure*}

In this section, we study the PNLMS and the IPNLMS algorithms aiming at verifying experimentally the properties and issues addressed 
in Section~\ref{sec:properties}.
We also consider two other algorithms: the NLMS serves as a benchmark, whereas the $\ell_0$-NLMS is used to illustrate the behavior of 
an algorithm that explicitly models the sparsity~\cite{Gu_l0_LMS_SPletter2009,Markus_sparseSMAP_tsp2014}. 
These four algorithms are used to identify some unknown systems/vectors $\wbf_{\rm o}$.

The parameters of the PNLMS and IPNLMS algorithms were set as explained in Section~\ref{sec:pt_nlms}. 
For the $\ell_0$-NLMS algorithm, we used the Laplacian function, and we set $\kappa=2 \times 10^{-3}$ and $\beta=5$, as in~\cite{Gu_l0_LMS_SPletter2009}.
For all algorithms, we set $\delta=10^{-12}$ and the step size $\mu$ is informed later.
The input signal is a zero-mean white Gaussian noise (WGN) with unit variance; the only exception is the input signal used in Fig.~\ref{fig:supmse}.
The measurement noise is also zero-mean WGN with variance $\sigma_n^2 = 10^{-2}$.
In each simulation scenario, the adaptive filter and the unknown systems have the same number of coefficients.
The MSE learning curves for each algorithm is generated by averaging the outcomes of $1000$ independent trials.

In Fig.~\ref{fig:supmse}, the unknown system is $\wbf_{{\rm o}}^{(0)}=[0.99~0.01]^T$.
Here, all algorithms use $\mu=0.4$, and we use a zero-mean binary phase-shift keying (BPSK) with unit variance 
as input signal\footnote{We opted for the BPSK input, rather than the WGN one, only for the sake of clarity of Fig.~\ref{fig:conv_path}, 
to avoid lines from crossing each other very often.}.
Figs.~\ref{fig:supmse1} and~\ref{fig:supmse2} depict the contours of the MSE surface for the NLMS and PNLMS algorithms, respectively. 
As expected, circular contours are observed in Fig.~\ref{fig:supmse1} since $\Rbf = \Ibf$ (uncorrelated input), whereas the 
contours in Fig.~\ref{fig:supmse2} are almost parallel lines since $\Rbf'= \Gbf_{\rm o}^{\frac{1}{2}} \Rbf \Gbf_{\rm o}^{\frac{1}{2}}$ 
is ill-conditioned. 
Indeed, while $\kappa(\Rbf)=1$, we have $\kappa(\Rbf')=\kappa(\Gbf_{\rm o})=99$.
These results corroborate the discussion in Subsection~\ref{sub:surface_MSE}. 
Moreover, the gradient direction corresponding to the contours in Fig.~\ref{fig:supmse2} is almost horizontal, explaining why the PNLMS algorithm 
is so fast to learn the high magnitude coefficient (horizontal), but very slow to learn the low magnitude coefficient (vertical), 
as illustrated in Fig.~\ref{fig:conv_path} and explained in Section~\ref{sec:properties}.

In Figs.~\ref{fig:bad_init00} and~\ref{fig:bad_init01}, the sparse unknown system is $\wbf_{\rm o}^{(1)} = [1~0~\dots~0]^T \in \Rset^{64}$,  
and the algorithms were initialized in two different ways: $\wbf(0) = {\bf 0}$ and $\wbf(0)=[0.05~\dots~0.05]^T$.
The step sizes for the NLMS, PNLMS, IPNLMS, and $\ell_0$-NLMS algorithms were set as $0.4$,~$0.3$,~$0.4$,~and~$0.99$, respectively, so that they 
could reach the same steady-state MSE. 
While in Fig.~\ref{fig:bad_init00} the PNLMS and IPNLMS algorithms converged as fast as the $\ell_0$-NLMS algorithm, their convergences were severely degraded  
by an initialization $\wbf(0)$ not matching the low magnitude coefficients of $\wbf_{\rm o}^{(1)}$, as illustrated in Fig.~\ref{fig:bad_init01}, 
confirming Remark~\ref{rem:initialization}.
Interestingly, the PNLMS and IPNLMS algorithms performed similar or worse than the NLMS algorithm, i.e., they did not succeed in exploiting the 
high sparsity degree in $\wbf_{\rm o}^{(1)}$, although $0.05$ is quite close to $0$.
This {fact} also explains the worse performance of proportionate-type algorithms in the more practical case of compressible (rather than sparse) vectors $\wbf_{\rm o}$, 
as can be verified in~\cite{Markus_sparseSMAP_tsp2014}.
Also, observe that the $\ell_0$-NLMS algorithm was able to exploit the system sparsity regardless of its initialization $\wbf(0)$.



In Fig.~\ref{fig:track}, we consider a non-stationary scenario in which the optimal coefficients are given by $\wbf_{\rm o}^{(2)}$ during 
the first 2000 iterations, and then they change to $\wbf_{\rm o}^{(3)}$, defined as
\begin{align} \label{eq:non_stationary}
\hspace{-3mm}
	\wbf_{{\rm o},i}^{(2)}    \!\!=\!\!
	\begin{cases}
	0,~\textrm{for}~0  \! \leq \! i  \! \leq \! 93,\\
	1,~\textrm{for}~94 \! \leq \! i  \! \leq \! 99,   
	\end{cases}   
	\! \wbf_{{\rm o},i}^{(3)}    \!\!=\!\!
	\begin{cases}
	1,~\textrm{for}~0 \! \leq \! i  \! \leq \! 5,\\
	0,~\textrm{for}~6 \! \leq \! i  \! \leq \! 99.   
	\end{cases}
\end{align}
{Both $\wbf_{\rm o}^{(2)}$ and $\wbf_{\rm o}^{(3)}$ have 6 nonzero coefficients out of 100.}
The algorithms were initialized with $\wbf(0) = {\bf 0}$, and the step sizes of the NLMS, PNLMS, IPNLMS, and $\ell_0$-NLMS algorithms were 
set as $0.6$, $0.15$, $0.45$, and $0.99$, respectively, so that they could achieve the same steady-state MSE. 
In the first 2000 iterations in Fig.~\ref{fig:track}, the PNLMS and IPNLMS algorithms were not as fast as the $\ell_0$-NLMS algorithm, although their coefficients 
were initialized properly and the unknown system is very sparse. 
This {behavior} is due to the support of $\wbf_{\rm o}^{(2)}$, as explained in Remark~\ref{rem:support}.
Besides, these Pt-NLMS algorithms were quite slow in the tracking of $\wbf_{\rm o}^{(3)}$, as illustrated in the last 2000 iterations in 
Fig.~\ref{fig:track} and described in Remark~\ref{rem:tracking}.


\section{Conclusion}
\label{sec:conclusion}

In this paper, we addressed the limitations of proportionate-type algorithms.
Although we focused on NLMS-based algorithms, the conclusions are valid for the whole family of proportionate algorithms, as 
they are due to the {\it proportional updates}, which accelerate the convergence of the high magnitude coefficients at the cost  
of slowing the convergence corresponding to the low magnitude coefficients.
The limitations together with the simulation results allow us to conclude that proportionate-type algorithms:
(i) do not really exploit sparsity; 
(ii) are not recommended for the tracking of non-stationary sparse systems whose support may change {during} the iterations; and 
(iii) may have a severe performance degradation when facing compressible systems.


\bibliographystyle{IEEEbib}
\bibliography{markus,adaptive_filtering,audio_acoustic,general_v2}

\balance


\end{document}